\documentclass[conference]{IEEEtran}
\IEEEoverridecommandlockouts

\usepackage{cite}
\usepackage{amsmath,amssymb,amsfonts}
\usepackage{algorithm}
\usepackage{algorithmic}
\usepackage{graphicx}
\usepackage{array}
\usepackage{subcaption}
\usepackage{textcomp}
\usepackage{xcolor}
\usepackage{tikz}
\usepackage{multirow}
\usepackage{soul}
\usetikzlibrary{arrows.meta,positioning,calc}
\usepackage{pgfplots}
\pgfplotsset{compat=1.18}

\def\BibTeX{{\rm B\kern-.05em{\sc i\kern-.025em b}\kern-.08em
    T\kern-.1667em\lower.7ex\hbox{E}\kern-.125emX}}

\makeatletter
\newcommand{\linebreakand}{%
    \end{@IEEEauthorhalign}
    \hfill\mbox{}\par
    \mbox{}\hfill\begin{@IEEEauthorhalign}
}
\makeatother
\begin{document}

\title{Toward a Foundation Plug-and-Play Prior for Computed Tomography Reconstruction via a Multimodal Diffusion Model}

\author{
\IEEEauthorblockN{Haley Duba-Sullivan}
\IEEEauthorblockA{
\textit{Oak Ridge National Laboratory}\\
Oak Ridge, TN, USA\\
sullivanhe@ornl.gov}
\and
\IEEEauthorblockN{Patxi Fernandez-Zelaia}
\IEEEauthorblockA{
\textit{Oak Ridge National Laboratory}\\
Oak Ridge, TN, USA\\
fernandezzep@ornl.gov}
\linebreakand
\IEEEauthorblockN{Obaidullah Rahman}
\IEEEauthorblockA{
\textit{Oak Ridge National Laboratory}\\
Oak Ridge, TN, USA\\
rahmano@ornl.gov}
\and
\IEEEauthorblockN{Amirkoushyar Ziabari}
\IEEEauthorblockA{
\textit{Oak Ridge National Laboratory}\\
Oak Ridge, TN, USA\\
ziabariak@ornl.gov}
\thanks{This manuscript has been authored by UT-Battelle, LLC, under contract DE-AC05-00OR22725 with the US Department of Energy (DOE). The US government retains and the publisher, by accepting the article for publication, acknowledges that the US government retains a nonexclusive, paid-up, irrevocable, worldwide license to publish or reproduce the published form of this manuscript, or allow others to do so, for US government purposes. DOE will provide public access to these results of federally sponsored research in accordance with the DOE Public Access Plan (https://www.energy.gov/doe-public-access-plan).}
}

\maketitle

\begin{abstract}
Computed tomography (CT) throughput is limited by scan time, which grows with both the number of projections acquired and the detector integration time for each projection.
Reconstructing high-quality volumes from sparse-view or low-dose measurements therefore depends on using an informative prior, typically a neural network trained for one specific scan setting and retrained whenever the modality, geometry, or material changes.
We investigate whether a single diffusion model trained across several imaging domains can instead serve as a reusable prior for heterogeneous CT reconstruction problems.
We evaluate the proposed method using the same diffusion visual transformer model and normalized denoising strength on three datasets that differ in modality, beam geometry, material, and degradation type, spanning additively manufactured metal parts and concrete microstructure imaged with cone-bean X-ray CT and parallel-beam neutron CT respectively.
The proposed method improves upon analytic reconstructions in all three cases, demonstrating transferability across the evaluated problems and providing a step toward a reusable foundation prior for heterogeneous CT reconstruction.
\end{abstract}


\section{Introduction}

\IEEEPARstart{C}{omputed} tomography (CT) is a critical tool for non-destructive characterization, from flaw and porosity analysis in additively manufactured (AM) metal parts to neutron characterization of irradiated concrete~\cite{ziabari_2023,duba-sullivan_2026,cheniour2024mesoscale}.
The quality of projections, however, is bounded by acquisition budget as much as by physics.
Scan time depends on both the number of projections acquired and the detector integration time spent on each one.
These two factors can be reduced independently.
Acquiring fewer projections leaves the reconstruction problem ill-posed, creating streaking artifacts in reconstructions from analytic methods such as filtered backprojection (FBP)~\cite{shepp_1974} and Feldkamp-Davis-Kress (FDK)~\cite{feldkamp_1984}.
Shortening the integration time decreases the number of photons recorded in each projection, which reduces the signal-to-noise ratio of the measured data.
High-throughput characterization depends on reconstruction methods that are accurate under both sparse-view and low-dose measurements.

Model-based reconstruction methods address this ill-posedness by introducing strong prior models paired with a data-fidelity term~\cite{venkatakrishnan_2013,chan_2016,kamilov_2023,duba-sullivan_2025}.
Recent work has shown that diffusion models are particularly strong priors~\cite{graikos_2022,wu_2024,park_2025,zhu_2023a,efimov_2026}. 
However, these learned priors are often trained for one specific CT scan setting (e.g., modality, scan geometry, and object material), and reconstructions generally degrade significantly when the scan settings change.
A common solution is to retrain a prior for each new problem or adapt the prior to the new distribution at test time~\cite{chung_2024,barbano_2025}.
Both of these options are too expensive for practical use at characterization facilities that must image many sample types across several instruments.

In this paper, we work toward a foundation prior model that can be used across a wide variety of CT scan settings without retraining or fine-tuning.
Specifically, we use a diffusion vision transformer (ViT) trained across eight imaging domains, ranging from metal AM parts to cell microscopy scans, as a shared denoiser within a plug-and-play (PnP) reconstruction framework.
We evaluate this prior on three distinct CT datasets that differ in modality (X-ray and neutron), scan geometry (cone- and parallel-beam), material (nickel and steel AM parts, and concrete microstructure), and degradation type (sparse view and short integration time).
The same frozen model weights, normalized denoising strength, and iteration count are used for every reconstruction, while the data-fidelity term accounts for the acquisition physics of each problem.

The main contributions of this work are as follows:
\begin{enumerate}
    \item We investigate a multimodal diffusion model as a reusable PnP prior for CT reconstruction, using a single denoising step per iteration without per-dataset retraining or test-time adaptation.

    \item We demonstrate the transferability of a single frozen prior across three heterogeneous CT reconstruction problems spanning different imaging modalities, scan geometries, materials, and degradation types.

    \item We show that a normalized denoising strength selected on one dataset transfers to the remaining datasets with at most a 0.5 dB reduction in PSNR relative to per-dataset tuning.
\end{enumerate}
Since these factors vary together across the three problems, our experiments establish transfer under their combined change rather than isolating the effect of each factor.

\section{Related Work}
\label{sec:related}
PnP reconstruction methods~\cite{venkatakrishnan_2013} replace the prior's proximal operator with a denoiser in an iterative reconstruction algorithm such as the alternating direction method of multipliers (ADMM)~\cite{boyd_2011a}.
Since the denoiser is applied independently of the forward model, the same prior can be paired with any data-fitting agent.
Early PnP methods used off-the-shelf denoisers such as BM3D~\cite{dabov_2007}, while later work replaced these with deep denoisers, either trained as general-purpose Gaussian denoisers or tailored to a specific imaging task~\cite{zhang_2021,ryu_2019,duba-sullivan_2025}.
PnP has since become a standard tool in computational imaging, with applications across many modalities~\cite{kamilov_2023}.

Diffusion models learn to reverse a forward process that progressively corrupts data with Gaussian noise~\cite{ho2020denoising,song_2020}.
During diffusion model training, weights of a network learn to predict the noise present in a corrupted input across a continuum of noise levels, yielding a denoiser parameterized by noise level~\cite{nichol_2021,karras_2022}.
Several recent works exploit this for CT reconstruction.
For example, Song et al.~\cite{song_2021} sample from a score-based prior at test time, Zhu et al.~\cite{zhu_2023a} fold a data-consistency step into diffusion sampling, Xia et al.~\cite{xia_2023a} regularize low-dose CT with a full denoising diffusion probabilistic model (DDPM) trajectory accelerated by Nesterov momentum, Chung et al.~\cite{chung_2023} apply a 2D diffusion prior slice-wise to 3D reconstruction with a total-variation penalty along the slice axis, and Hossain et al.~\cite{hossain_2026} regularize an implicit neural representation with a diffusion prior for self-supervised neutron CT reconstruction.

The cost of running the full reverse diffusion process conditioned on the measurements is a significant limitation for iterative CT reconstruction, where a single volume spans on the order of a thousand slices and the prior is evaluated for each slice at every outer iteration.
Park et al.~\cite{park_2025} interpret PnP as a score-based method, showing that a pretrained diffusion model can be substituted into a classical PnP algorithm without invoking the reverse stochastic differential equation that generative sampling requires; Wu et al.~\cite{wu_2024} reach a related construction from a Bayesian direction, reducing posterior sampling to a sequence of Gaussian denoising problems.
Both use the diffusion model as a one-step denoiser inside an optimization loop.
In this work, we adopt the same interpretation.

A pretrained diffusion prior may not transfer reliably when the test distribution differs from its training distribution, and reconstruction quality can degrade under such distribution shifts~\cite{thomsen_2026,zheng_2025}.
One response is to adapt the prior at test time, either by casting adaptation in terms of the deep image prior~\cite{chung_2024} or by steering the sampling trajectory using the measurements~\cite{barbano_2025}.
Such adaptation can recover accuracy, but adds per-problem computation at test time and therefore limits throughput.
An alternative is to broaden the training distribution so that a single model can support many imaging problems~\cite{terris_2025a,xia_2026}.
We follow this direction while retaining the PnP formulation.
Rather than train an end-to-end reconstruction model, we use a single broadly trained diffusion model as the shared denoiser, leaving the acquisition physics to the forward operator.
This separation allows the same frozen prior to be applied across scan settings without retraining or adapting at test-time.

\section{Methodology}
\label{sec:method}

\subsection{Problem Formulation}
The goal in CT reconstruction is to recover a 3D volume of linear attenuation coefficients $x$ from noisy measured projections $y$ by inverting the measurement process.
Model-based reconstruction methods estimate $\hat{x}$ using a regularized weighted least-squares formulation, i.e.
\begin{equation}
\hat{x} = \arg \min_x \left\{\tfrac{1}{2} \|y-Ax\|^2_{\Lambda} + g(x)\right\},
\label{eq:mbir}
\end{equation}
where $A$ is the geometry- and modality-specific forward operator, $\Lambda$ is a diagonal weighting matrix derived from the measurement statistics, and $g(x)$ is a regularization function.

\subsection{PnP-ADMM}

Variable splitting recasts \eqref{eq:mbir} as a constrained problem that decouples the data-fidelity and regularization terms, tied together by an equality constraint.
ADMM~\cite{boyd_2011a} solves the constrained problem by alternating a data-fidelity update, a prior update, and a dual update.
Letting $f(x) = \tfrac{1}{2} \|y-Ax\|^2_{\Lambda}$ be the data-fidelity term, these subproblems are given by
\begin{align}
v_{k+1} &= \arg\min_v \left\{f(v) + \tfrac{1}{2\gamma^2}\|v - (x_{k} - u_{k})\|^2 \right\}, \label{eq:vupdate}\\
x_{k+1} &= \arg\min_x \left\{ g(x) + \tfrac{1}{2\gamma^2}\|x - (v_{k+1} + u_{k})\|^2 \right\}, \label{eq:xupdate}\\
u_{k+1} &= u_{k} + v_{k+1} - x_{k+1}, \label{eq:uupdate}
\end{align}
where $u$ is the scaled dual variable and $\gamma>0$ is a step-size parameter.

The data-fidelity subproblem given by~\eqref{eq:vupdate} is a quadratic problem determined entirely by the acquisition physics.
Its solution satisfies the normal equations
\begin{equation}
\big(A^{\!\top}\!\Lambda A + \gamma^{-2} I\big)\, v_{k+1} = A^{\!\top}\!\Lambda y + \gamma^{-2}\big(x_k - u_k\big),
\label{eq:normaleq}
\end{equation}
which we solve iteratively with conjugate gradient, warm-started from the previous outer iterate.

The regularization problem given by~\eqref{eq:xupdate} is exactly the MAP estimate for removing additive white Gaussian noise of standard deviation $\gamma$ under the prior implied by $g$, as originally observed in~\cite{venkatakrishnan_2013}.
So, this subproblem can be replaced with a denoiser $D_\theta(\cdot, \gamma)$ evaluated at noise level $\gamma$, i.e.
\begin{equation}
x_{k+1} = D_\theta(v_{k+1} + u_k;\, \gamma).
\label{eq:pnpsub}
\end{equation}
Note that this subproblem does not involve the measurements or forward operator, calling into question whether a single $D_\theta$ can serve across a variety of scan settings.

\subsection{Multimodal Diffusion Model}
\label{sec:prior}

We use a diffusion model with a ViT backbone~\cite{peebles_2022,ho2020denoising,dosovitskiy2020image} that we train across several imaging domains.
The resulting noise-prediction network $\epsilon_\theta(\cdot \mid t)$ predicts the noise in an input drawn from the model's training marginal at time step $t$, so it is not itself the denoiser required by \eqref{eq:pnpsub}, which must accept an arbitrary noise standard deviation $\gamma>0$.
To reconcile this, we construct the denoiser $D_\theta$ from $\epsilon_\theta$ using a deterministic denoising diffusion implicit model (DDIM) reverse map~\cite{song_2020a}, which requires resolving two mismatches.
First, $\gamma$ carries the physical units of the scan, whereas the model's noise schedule is defined on its own normalized intensity scale.
We therefore rescale the input by an intensity scale taken from the initial reconstruction, $s = \mathrm{std}(x_0)$, and express the requested noise level in model space as the dimensionless quantity $\sigma = \gamma / s$.
We stress that $s$ is a scale statistic of the initial reconstruction; it is dominated by object contrast rather than by the noise and serves only to place different scans on a common intensity scale.
Accordingly, $\sigma$ is a normalized denoising strength rather than a physical noise level.
Second, $\epsilon_\theta$ is defined only at the discrete time steps of the training schedule, so given a noisy input $z = x + \gamma\epsilon$ with $\epsilon \sim \mathcal{N}(0, I)$ we select the step whose training noise level is closest to $\sigma$, i.e.
\begin{equation}
t^* = \arg\min_t \left|\, \sigma - \sqrt{(1-\bar\alpha_{t})/\bar\alpha_{t}} \,\right|,
\label{eq:tstar}
\end{equation}
where $\bar\alpha_t = \prod_{i=1}^{t}\alpha_i$ is the model's cumulative noise schedule.
Writing $\hat z = z/s$ for the rescaled input and $\hat\epsilon_{t^*} = \epsilon_\theta(\sqrt{\bar\alpha_{t^*}}\, \hat z \mid t^*)$ for a single noise prediction, the reverse map gives
\begin{equation}
D_\theta(z;\gamma)
= s\,\frac{\sqrt{\bar\alpha_{t^*}}\, \hat z  - \sqrt{1-\bar\alpha_{t^*}}\,\hat\epsilon_{t^*}}{\sqrt{\bar\alpha_{t^*}}}
= z - \gamma\,\hat\epsilon_{t^*} ,
\label{eq:onestep}
\end{equation}
where the second equality holds up to the discretization of the noise schedule, since $t^*$ is chosen in \eqref{eq:tstar} so that $\sqrt{(1-\bar\alpha_{t^*})/\bar\alpha_{t^*}}$ is the closest available training noise level to $\sigma$.
Evaluating the prior requires only a single diffusion time step per PnP iteration, rather than a trajectory over the schedule, so its per-iteration cost matches that of a conventional PnP denoiser rather than that of a diffusion sampler.

We use the same model weights for every reconstruction reported in this paper, with no fine-tuning, test-time adaptation, or per-dataset retraining.
The denoiser uses a ViT backbone with five transformer encoder layers, eight attention heads, a latent dimension of 512, a patch size of $4\times4$, fixed sinusoidal positional embeddings, and a linear decoder head.
The network accepts $128\times128$ images and contains $39.5$ million trainable parameters.

\begin{figure}[htbp]
    \begin{center}
        \includegraphics[width=1.0\linewidth]{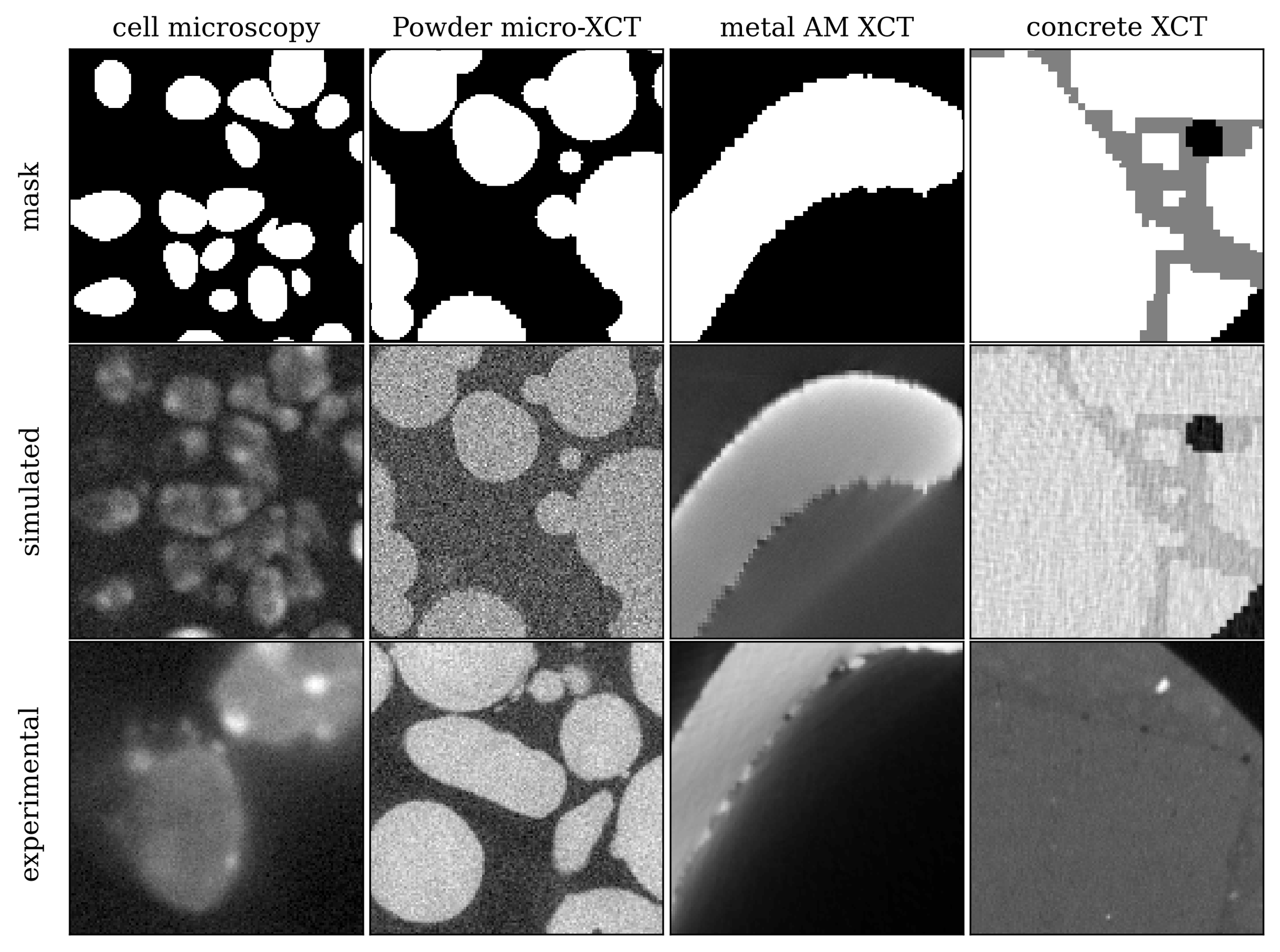}
        \caption{Diffusion model training data: cell microscopy, powder micro-XCT, metal AM XCT, and concrete XCT. Top row: simulation masks. Middle row: simulated response. Bottom row: unpaired experimental examples.}
        \label{fig:trainingExamples}
    \end{center}
\end{figure}

We train the diffusion model on four imaging modalities, each containing experimental and simulated data and therefore yielding eight distinct imaging domains.
The training set contains metal AM turbine-blade XCT \((N=54{,}238)\)~\cite{ziabari_2023,ziabari2020beam}, biological cell microscopy \((N=28{,}000)\)~\cite{ziabari2024yolo2u}, metal powder micro-XCT \((N=24{,}498)\)~\cite{ziabari2022system}, and irradiated concrete XCT \((N=46{,}230)\)~\cite{cheniour2024mesoscale}.
Each modality contains an equal mixture of experimental measurements and simulated examples.
Figure~\ref{fig:trainingExamples} shows representative training samples.
We standardize the data using statistics estimated from the corresponding training set.

We use a linear diffusion schedule with \(T=1000\) steps, \(\beta_{t=1}=10^{-4}\), and \(\beta_{t=T}=0.02\). 
During training, the model receives two forms of conditioning.
A one-hot domain label identifies the imaging domain, such as cell-real or powder-simulated, while simulated examples can additionally provide spatial masks identifying the material phases used by the physics-based simulation.
We provide conditioning with probability \(p_{\mathrm{cond}}=0.50\), allowing the model to operate both conditionally and unconditionally.
For all PnP reconstructions in this work, we evaluate the denoiser unconditionally.

We train the model for 650 epochs on an NVIDIA DGX system with eight H100 80 GB GPUs using the simplified noise-prediction objective of~\cite{ho2020denoising}.
We use Adam with a learning rate of \(10^{-5}\) and a batch size of 256.
On-the-fly augmentation consists of random \(90^\circ\) rotations and horizontal and vertical flips.

The prior operates on 2D images while the reconstruction is a 3D volume, so we apply the denoiser independently to each axial slice and restack the outputs.
Since the network accepts fixed $128\times128$ inputs, we cover each slice with overlapping patches and recombine their outputs using squared-sine window averaging.
Although the data-fidelity update operates on the full 3D volume, the slice-wise denoiser does not explicitly enforce consistency between adjacent slices.

 \section{Implementation Details}

\begin{table*}
\fontsize{8}{11}
\selectfont
\centering
\caption{Overview of three testing datasets, which differ in imaging modality, beam geometry, and material. Bracketed entries give the input and reference acquisitions respectively. Both X-ray datasets are beam hardening and scatter corrected before reconstruction. The cone-beam short scans span $180^\circ$ plus the $17^\circ$ fan angle, while the parallel-beam neutron scan requires only $180^\circ$. Note that the nickel data is simulated, so it has no detector integration time; the equivalent photon budget is set by the noise added to its projections instead.}
\label{tab:datasets}
\begin{tabular}{|l|ccc|}
\hline
 & Nickel AM & Steel AM & Concrete Microstructure \\ \hline
Modality & X-ray & X-ray & Neutron \\
Geometry & Cone-beam & Cone-beam & Parallel-beam \\
Data source & Simulated & Measured & Measured \\
Angle range [Input, Ref.] & [$197^\circ$, $360^\circ$] & [$197^\circ$, $197^\circ$] & [$180^\circ$, $180^\circ$] \\
\# of views [Input, Ref.] & [145, 2132] & [1000, 1000] & [34, 546] \\
Integration time [Input, Ref.] & -- & [0.6\,s, 3.6\,s] & [30\,s, 30\,s] \\
Reconstruction [Input, Ref.] & [FDK, FDK] & [FDK, MBIR] & [FBP, FBP] \\
Volume (slices $\times$ rows $\times$ cols) & 1224$\,\times\,$1122$\times$1295 & 1264$\times$1360$\times$1360 & 2024$\times$1048$\times$1048 \\
Voxel size ($\mu$m) & 17.28 & 17.28 & 50 \\
\hline
\end{tabular}
\end{table*}

\subsection{Datasets}
We evaluate the proposed method on three CT datasets, summarized in Table~\ref{tab:datasets}.
Together, these datasets span two imaging modalities (X-ray and neutron), two beam geometries (cone- and parallel-beam), three materials, and both simulated and measured acquisitions.

\textbf{Simulated Ni AM part (cone-beam X-ray CT).} 
We use a CAD model of an AM nickel part as a simulation phantom, providing a controlled evaluation with known object geometry.
Projections are generated using SpekPy~\cite{poludniowski_2021,bujila_2020} at 200 kV through a $1456\times1840$ CsI detector with 0.127 mm pitch.
The input short scan additionally carries added noise and simulated scatter, while the reference full scan has neither, so this dataset pairs sparse angular sampling with controlled noise and scatter degradation.
Both scans are polychromatic and are corrected for scatter and beam hardening using LEAP~\cite{kim_2023a} before reconstruction with FDK~\cite{feldkamp_1984}.
The reference is therefore not the CAD phantom itself, but a full-view, noise- and scatter-free acquisition carried through the same correction and reconstruction path as the input, which holds the beam physics fixed so that the metrics isolate the sparse-view and noise degradation.

\textbf{Measured steel AM part (cone-beam X-ray CT).} An AM 316L steel part is scanned at 200\,kV with a 2\,mm Al source filter.
Input and reference are both 1000-view short scans acquired with different detector integration times, giving total acquisition times of 10 and 60 minutes respectively.
The input is reconstructed with FDK and the reference with MBIR~\cite{thibault_2007}, both after scatter and beam hardening correction using LEAP~\cite{kim_2023a}.

\textbf{Concrete microstructure (parallel-beam neutron CT).} A concrete microstructure sample is imaged at the CG-1D beamline of the High Flux Isotope Reactor on a $2048\times2048$ detector, with a $7\times7$ median filter applied to the projection data to suppress salt-and-pepper noise in the measurement.
The scan spans $180^\circ$ with 546 projections; the input subsamples 34 of these, corresponding to roughly a $16\times$ reduction in angular sampling.
Unlike the other two datasets, the degradation here is purely angular, and both input and reference reconstructions use FBP.
This is also the most out-of-distribution data relative to the prior's predominantly X-ray training data.

\subsection{Baselines and Metrics}
\label{sec:metrics}

We compare the proposed PnP-Diffusion-ViT method against the analytic input reconstruction and PnP-BM3D~\cite{dabov_2007}.
PnP-BM3D uses the same PnP-ADMM algorithm, iteration count, and nominal denoising strength as the proposed method, with the BM3D noise level obtained using the same intensity normalization.

Metrics are computed against the reference reconstruction over a masked center sub-volume of 16 slices.
The mask is obtained by Otsu thresholding the reference and filling interior holes so that internal voids and flaws remain included in the evaluation.
We report the peak signal-to-noise ratio (PSNR), structural similarity index measure (SSIM)~\cite{wang_2004}, and high-frequency error norm (HFEN)~\cite{ravishankar_2011}, where higher PSNR and SSIM indicate better agreement with the reference and lower HFEN indicates better high-frequency fidelity.
HFEN measures the reconstruction error after Laplacian-of-Gaussian filtering, making it particularly sensitive to discrepancies in edges, fine structures, and other high-frequency features.
PSNR and SSIM can favor over-smoothed reconstructions, so HFEN complements these metrics by penalizing the loss of high-frequency detail.

\subsection{Algorithmic Parameters}
The physical step size $\gamma$ depends strongly on the scale of each reconstruction, with its optimal value differing by two orders of magnitude across the three datasets.
We therefore parameterize the algorithm using the dimensionless normalized denoising strength $\sigma=\gamma/s$ introduced in Section~\ref{sec:prior}.
We select $\sigma=0.25$ by a parameter sweep on the nickel AM dataset and use this value unchanged for the steel and concrete datasets.
All reconstructions use $5$ PnP iterations.
Thus, the denoising strength and iteration count remain fixed as the imaging modality, geometry, material, and degradation type change.

To evaluate how well the selected denoising strength transfers, we also sweep $\sigma$ independently on each dataset.
The optimal values are $\sigma^\star=0.25$ for nickel and $\sigma^\star=0.29$ for both steel and concrete.
The shared value $\sigma=0.25$ therefore remains close to the independently optimized value for all three datasets, even though the corresponding physical step sizes $\gamma$ differ by two orders of magnitude.
Using the shared value instead of the per-dataset optimum reduces PSNR by at most $0.5$\,dB.

\section{Experimental Results}
\label{sec:results}

\begin{figure*}
	\centering
	\begin{subfigure}{\textwidth}
		\centering
		\includegraphics[width=0.95\textwidth]{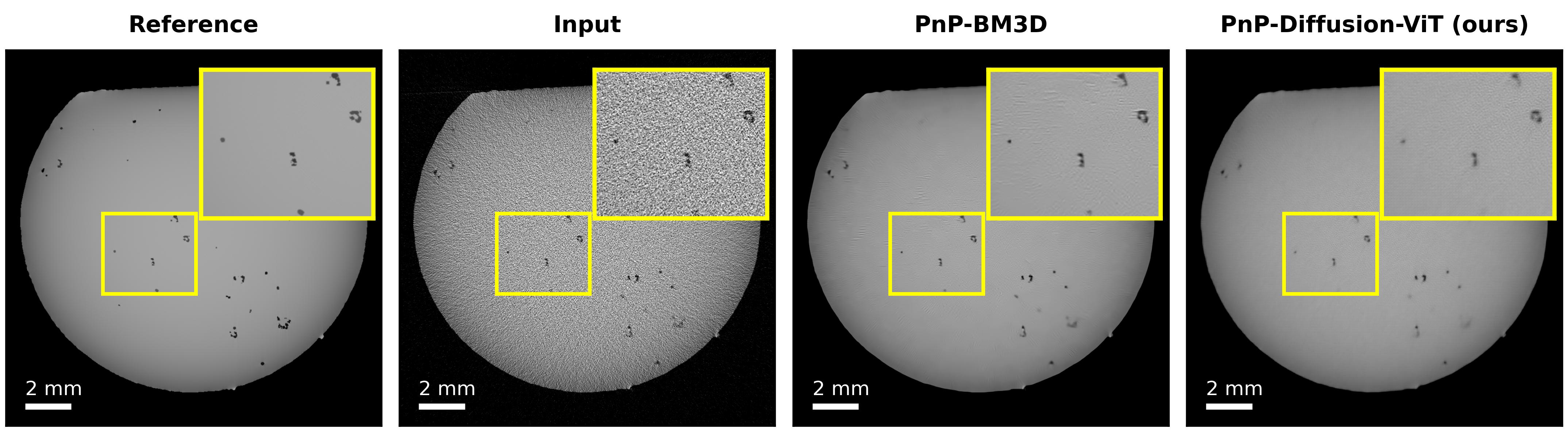}
		\caption{Simulated nickel AM part (cone-beam X-ray CT)}
		\label{fig:qual-nickel}
	\end{subfigure}\\
	\begin{subfigure}{\textwidth}
		\centering
		\includegraphics[width=0.95\textwidth]{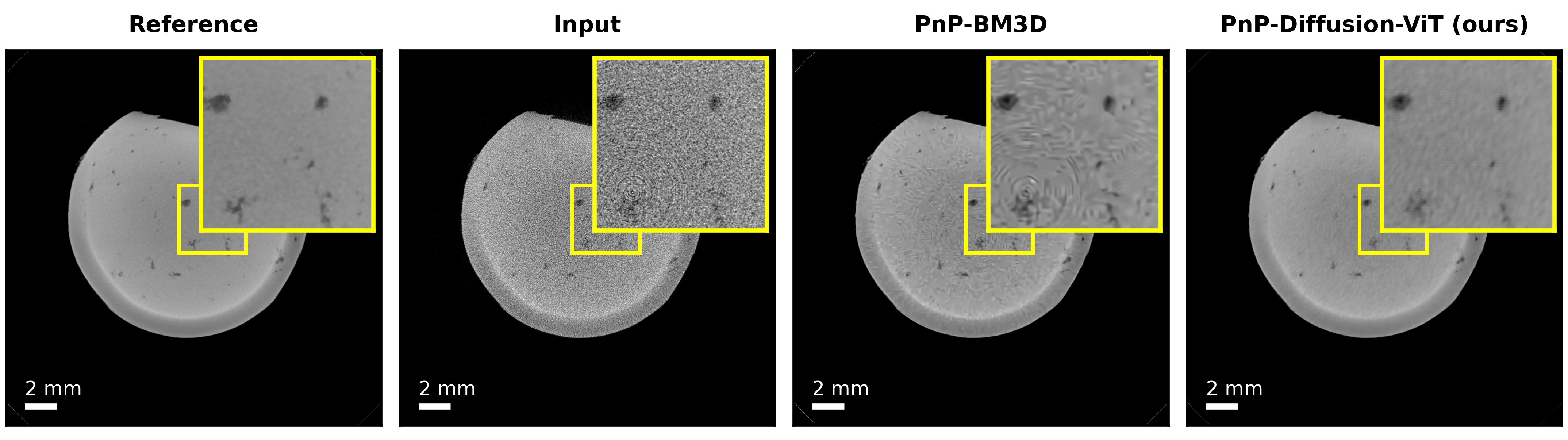}
		\caption{Measured steel AM part (cone-beam X-ray CT)}
		\label{fig:qual-steel}
	\end{subfigure}\\
	\begin{subfigure}{\textwidth}
		\centering
		\includegraphics[width=0.95\textwidth]{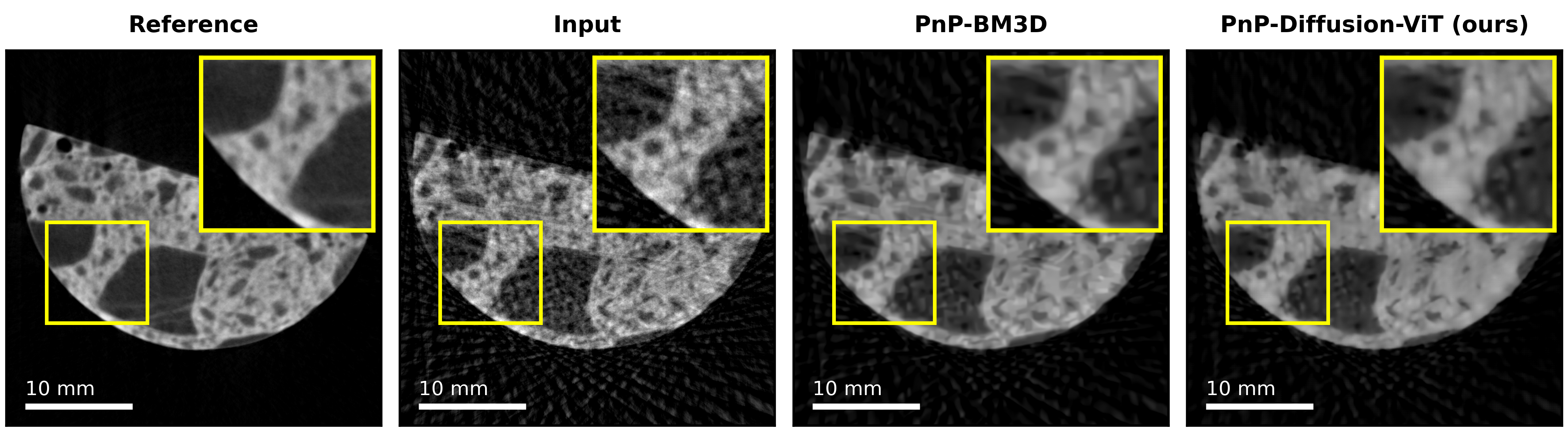}
		\caption{Measured concrete microstructure (parallel-beam neutron CT)}
		\label{fig:qual-concrete}
	\end{subfigure}
	\caption{Qualitative comparison for each dataset. Columns from left to right show the reference reconstruction, the analytic input, the PnP-BM3D reconstruction, and the proposed PnP-Diffusion-ViT reconstruction. The reference reconstructions are generated using full-view FDK for nickel AM, long-integration-time MBIR for steel AM, and full-view FBP for concrete microstructure. Both PnP reconstructions use the same algorithmic parameters ($\sigma=0.25$, $K=5$). The same frozen prior removes the dominant degradation of all three problems (noise, ring artifacts, and sparse-view streaking) while retaining fine structure that BM3D smooths away.}
	\label{fig:qualitative}
\end{figure*}

Figure~\ref{fig:qualitative} compares the reference reconstruction, analytic input, PnP-BM3D, and the proposed PnP-Diffusion-ViT reconstruction for each dataset.
For the nickel AM part, the analytic input contains strong high-frequency noise that obscures the interior flaw structure.
Both PnP methods suppress this noise and recover the flaws visible in the reference.
For the steel AM part, the input contains both noise and a prominent ring artifact.
PnP-Diffusion-ViT suppresses both while preserving the small interior flaws, whereas residual structured artifacts remain in the PnP-BM3D reconstruction.
For the concrete sample, sparse angular sampling produces severe streaking that both PnP methods substantially reduce.
The PnP reconstructions are smoother than the reference, particularly for PnP-BM3D, consistent with the HFEN values reported in Table~\ref{tab:results}.
The remaining degradation is most apparent for the concrete data, which differs most strongly from the training distribution in both imaging modality and material.

\begin{table}
	\caption{Quantitative comparison against the reference reconstruction. Best metric per dataset is bolded; higher is better for PSNR and SSIM, and lower is better for HFEN. Time is reported in seconds per slice, excluding data loading and output writing.}
	\label{tab:results}
	\begin{center}
		\footnotesize
        \setlength{\tabcolsep}{4pt}
        \renewcommand{\arraystretch}{1.15}
        \begin{tabular}{|c|c|r|r|r|c|}
			\hline
			\textbf{Dataset} & \textbf{Method} & \textbf{PSNR} & \textbf{SSIM} & \textbf{HFEN} & \textbf{Time (s\ /\ slice)} \\
			\hline
			\multirow{3}{*}{\shortstack[l]{Nickel AM}}
            & FDK & 9.446 & 0.026 & 3.710 & 0.002 \\
            & PnP-BM3D & \textbf{25.769} & \textbf{0.823} & 0.744 & 40.8 \\
            & PnP-Diff-ViT & 25.292 & 0.737 & \textbf{0.733} & 3.67 \\
            \hline
			\multirow{3}{*}{\shortstack[l]{Steel AM}}
			& FDK & 13.650 & 0.060 & 3.331 & 0.008 \\
            & PnP-BM3D & 24.580 & 0.601 & 0.872 & 47.7 \\
            & PnP-Diff-ViT & \textbf{26.680} & \textbf{0.700} & \textbf{0.571} & 9.59 \\
            \hline
			\multirow{3}{*}{\shortstack[l]{Concrete}}
            & FBP & 14.872 & 0.256 & 1.796 & 0.001 \\
            & PnP-BM3D & 16.819 & 0.437 & 1.026 & 27.8 \\
            & PnP-Diff-ViT & \textbf{16.984} & \textbf{0.450} & \textbf{0.985} & 2.12 \\
            \hline
		\end{tabular}
	\end{center}
\end{table}

Table~\ref{tab:results} reports metrics with respect to the reference reconstruction. 
PnP-Diffusion-ViT substantially improves on the analytic input for all three datasets using the same frozen prior and denoising strength.
PSNR increases by $15.8$ dB for nickel, $13.0$ dB for steel, and $2.1$ dB for concrete, while HFEN decreases by $80\%$, $83\%$, and $45\%$, respectively.
These results demonstrate that the same frozen prior and normalized denoising strength can be reused across the three evaluated CT reconstruction problems, which differ simultaneously in imaging modality, beam geometry, material, and degradation type.
Relative to PnP-BM3D, PnP-Diffusion-ViT achieves the best PSNR, SSIM, and HFEN on both measured datasets.
It improves PSNR by $2.1$ dB on steel and $0.17$ dB on concrete.
For the simulated nickel data, PnP-BM3D achieves $0.48$ dB higher PSNR and $0.086$ higher SSIM, while PnP-Diffusion-ViT achieves slightly lower HFEN.
The stronger PSNR and SSIM of BM3D on nickel are consistent with the approximately additive white Gaussian noise used in its simulation, which closely matches the noise model assumed by BM3D.
PnP-Diffusion-ViT nevertheless achieves the lowest HFEN on all three datasets, indicating stronger preservation of high-frequency structure relative to the reference.

Table~\ref{tab:results} also reports reconstruction time in seconds per slice, excluding data loading and output writing.
PnP-Diffusion-ViT requires 2.1-9.6 s per slice using four NVIDIA H100 80 GB GPUs, whereas PnP-BM3D requires 27.8-47.7 s per slice using a CPU implementation on a system equipped with two AMD EPYC 9654 processors.
Since the methods are evaluated on different hardware, the reported runtimes characterize their respective implementations rather than establishing a hardware-matched speed advantage.
The proposed method requires only a single diffusion denoising step per PnP iteration, avoiding the computational cost of a full reverse diffusion trajectory.
For the $K=5$ reconstructions, this corresponds to five denoising passes per slice, with each pass evaluating the diffusion model on overlapping image patches.

\section{Conclusion}
\label{sec:conclusion}
In this paper, we demonstrated that a multimodal diffusion model trained across several imaging domains can serve as a PnP prior for CT reconstruction problems that differ in modality, beam geometry, material, and degradation type, without retraining or test-time adaptation.
Using a single deterministic DDIM step per PnP iteration avoids the cost of a full reverse diffusion trajectory while retaining the diffusion model as an effective denoiser.
Across all three datasets, the proposed Diffusion-ViT prior substantially improves upon the analytic reconstruction and achieves the lowest HFEN.
Moreover, a normalized denoising strength selected on the nickel dataset transfers unchanged to the steel and concrete datasets with at most a 0.5 dB loss in PSNR relative to per-dataset tuning, despite the corresponding physical step sizes differing by two orders of magnitude.

These results demonstrate the feasibility of reusing a single broadly trained diffusion prior across the three heterogeneous CT reconstruction problems considered in this work.
However, the present experiments do not isolate the contribution of multimodal training to this transferability, as no architecture-matched single-domain prior is evaluated.
Furthermore, the evaluation is limited to three datasets, some of which share material characteristics with the training distribution.
Additionally, because the prior is applied independently to 2D slices, the present experiments do not directly evaluate inter-slice consistency or the preservation of three-dimensional structural continuity.
Future work will compare multimodal and single-domain priors under matched training conditions, evaluate transfer to additional held-out modality and material families, and investigate approaches that explicitly incorporate inter-slice information.

\bibliographystyle{IEEEtran}
\bibliography{references}

\end{document}